\documentclass{article}
\usepackage[utf8]{inputenc}
\usepackage[T1]{fontenc}
\usepackage[english]{babel}
\usepackage[top=2.5cm, bottom=2.5cm, left=2.75cm, right=2.75cm, letterpaper]{geometry}
\usepackage{multicol}
\usepackage{graphicx}
\usepackage{wrapfig}
\usepackage{amsmath}
\usepackage{listings}
\usepackage{titling}
\usepackage{bm}
\usepackage{float}
\usepackage{color}
\usepackage{subcaption}
\usepackage{authblk}
\usepackage{siunitx}
\usepackage{soul} 
\usepackage[labelfont=bf]{caption}

\usepackage{upgreek}                 
\usepackage{indentfirst}        
\usepackage[mathb]{mathabx} 
\usepackage[cal=cm,bb=libus,bbscaled=1.07,frak=euler]{mathalpha} 

\graphicspath{ {./Figures/} } 

\usepackage{csquotes}

\usepackage{CJKutf8}

\usepackage{siunitx}
\DeclareSIUnit\molar{\textsc{M}}
\DeclareSIUnit\adu{ADU}
\DeclareSIUnit\electron{\mathrm{e^-}}
\DeclareSIUnit\photon{photons}

\usepackage{hyperref}
\usepackage[capitalise, noabbrev]{cleveref} 
\crefformat{equation}{#2Equation~#1#3} 
\crefformat{figure}{#2Figure~#1#3}           

\newcommand{\PSF}{\text{PSF}}

\newcommand{\prob}[1]{\mathcal{P}\left(#1\right)}

\newcommand{\poisson}[1]{\textbf{Poisson}\left(#1\right)}

\usepackage[dvipsnames]{xcolor} 

\begin{document}
\title{Re-thinking 
Richardson-Lucy without Iteration Cutoffs:  Physically Motivated Bayesian Deconvolution}

\author[1,2]{Zachary H. Hendrix}
\author[1,2]{Peter T. Brown}
\author[1,2]{Tim Flanagan}
\author[1,2]{Douglas P. Shepherd}
\author[1,2,*]{Ayush Saurabh}
\author[1,2,3,*]{Steve Press{\'e}}
\affil[1]{Center for Biological Physics, Arizona State University, Tempe, AZ, USA}
\affil[2]{Department of Physics, Arizona State University, Tempe, AZ, USA}
\affil[3]{School of Molecular Sciences, Arizona State University, Tempe, AZ, USA}
\affil[*]{spresse@asu.edu; asaurabh@asu.edu}
\date{\today}
\maketitle
\begin{abstract}
\noindent
Richardson-Lucy deconvolution is widely used to restore images from degradation caused by the broadening effects of a point spread function and corruption by photon shot noise, in order to recover an underlying object. In practice, this is achieved by iteratively maximizing a Poisson emission likelihood. However, the RL algorithm is known to prefer sparse solutions and overfit noise, leading to high-frequency artifacts. The structure of these artifacts is sensitive to the number of RL iterations, and this parameter is typically hand-tuned to achieve reasonable perceptual quality of the inferred object. Overfitting can be mitigated by introducing tunable regularizers or other {\it ad hoc} iteration cutoffs in the optimization as otherwise incorporating fully realistic models can introduce computational bottlenecks. To resolve these problems, we present Bayesian deconvolution, a rigorous deconvolution framework that combines a physically accurate image formation model avoiding the challenges inherent to the RL approach. Our approach  achieves deconvolution while satisfying the following desiderata: 

\begin{itemize}

\item[I] deconvolution is performed in the spatial domain (as opposed to the frequency domain) where all known noise sources are accurately modeled and integrated in the spirit of providing full probability distributions over the density of the putative object recovered;
\item[II]  the probability distribution is estimated without making assumptions on the sparsity or continuity of the underlying object;
\item[III]  unsupervised inference is performed and converges to a stable solution with no user-dependent parameter tuning or iteration cutoff;
\item[IV] deconvolution produces strictly positive solutions; and
\item[V] implementation is amenable to fast, parallelizable computation.
\end{itemize}
\end{abstract}

\section{Introduction}

Deconvolution improves contrast and recovers small, or equivalently, high spatial frequency features in images obtained by diffraction-limited optical equipment such as telescopes and microscopes. However, diffraction of light ultimately limits  spatial frequencies supported by the optics, leading to attenuation and loss of high-frequency features~\cite{goodman2005introduction}. All information from spatial frequencies above the microscope bandpass is lost in the image formation process making the inverse problem of recovering the underlying object ill-posed. The ill-posedness is exacerbated by the Poisson noise originating from the quantum nature of photons, the detector noise, and inherent data pixelization.  
In light of the stochasticity introduced by the image formation model, the full inverse problem is naturally re-pitched probabilistically where high-frequency features outside the bandpass set by the Fourier transform of the point spread function~(PSF) cannot be resolved with certainty. 

In the 1970s, William H. Richardson and Leon B. Lucy independently proposed a computationally inexpensive algorithm to restore diffraction-degraded images through an iterative process maximizing the observation likelihood, assuming Poisson noise only. The resulting algorithm is now known as Richardson-Lucy~(RL) deconvolution \cite{Richardson, Lucy}. While Poissonian noise is a reasonable approximation in the higher signal-to-noise ratio~(SNR) regime, with large photon counts, the true noise model deviates considerably when the detector noise is significant~\cite{mannam2022real}. Furthermore, maximum likelihood methods are prone to overfitting data and, in particular, the RL algorithm prefers sparse solutions and exhibits overfitting by amplifying noise and generating artifacts within an image~\cite{Bertero}. A contemporaneous alternative, maximum entropy deconvolution~\cite{Burg}, was later combined with RL to help identify an iteration stop criterion~\cite{Boerner}.

To keep the noise model simple and avoid extra parameters irrelevant to high SNR data, many modern approaches still only consider Poissonian noise and only improve upon the likelihood maximization problem by introducing regularization. Regularizers are added to the likelihood and a two-term convex optimization problem is then solved for a single point object estimate. While these approaches have been successful in many contexts, they often require hand-crafted regularizers with carefully tuned hyperparameters~ \cite{Meiniel2018, fan_brief_2019}. Most commonly used regularizers can be classified according to the types of structures they are designed to encourage to help reduce ill-posedness: smooth or sparse. Regularizers  such as total variation~(TV)~\cite{Rudin} and  Hessian~\cite{huang2018fast} encourage object smoothness while others such as compressed sensing encourage sparsity~\cite{eslahi2016compressive}. 

The TV regularizer is the most popular regularization technique and involves minimizing the total variation in the deconvolved image,  which encourages smoothness while preserving sharp edges. Recent implementations of TV have leveraged fast iterative shrinkage-thresholding algorithm~(FISTA)~\cite{beck2009fast} and alternating direction method of multipliers~(ADMM)~\cite{chan2013constrained}. However, these techniques often result in restored images manifesting staircase artifacts whereby reconstructed images exhibit piecewise constant structures and loss of fine details~({\it e.g.}, textures)~\cite{Rudin, Dey} visually adjusted via the regularization parameter. 

Hessian matrices are yet another regularizer encouraging smoothness but avoids oversharpening of boundaries between regions of significantly different intensities~\cite{huang2018fast}. Here, unlike TV penalty where the integral over the first order derivatives or pixel-to-pixel variation is minimized, integration is performed over second-order derivatives enforcing differentiability of the object at every point. However, while differentiability encourages smoothness of transitions between regions, it may result in loss of resolution.

On the other hand, compressed sensing~\cite{eslahi2016compressive} is an optimization technique that allows deconvolving or denoising images where the data is sparse in some basis~(for instance, the Fourier basis or a wavelet basis~\cite{blunck2020compressed}). A combination of sparsity and continuity assumptions have also been used in image restoration~\cite{zhao2022sparse}. Such strategies require advance knowledge of the underlying object of interest, which may not always be available.

More recently, neural networks have allowed fast deconvolutions without requiring any knowledge of the noise sources. On one hand, prior knowledge of diffraction or blurring physics can be incorporated implicitly by parameterizing the unblurred image using a specific network architecture. For example, using either convolutional neural networks (CNNs) as in the deep image prior (DIP) approach~\cite{ulyanov2018deep} or coordinate-based neural radiance fields as implicit neural representations~\cite{ma2022deblur} have been successful in various contexts. However, these networks are typically optimized iteratively using the input raw images only and have a tendency to overfit data without a suitable stopping criterion~\cite{ulyanov2018deep, ma2022deblur}. On the other hand, many fully supervised methods~\cite{ni2024pi, qiao2024zero,xu2014deep, yanny2022deep, li2022incorporating} have been developed demanding large amounts of input network training data before deconvolution can be performed and raising concerns as to their generalizability in deconvolving different structures.

\begin{figure}[H]
    \centering
\includegraphics[width=\linewidth]{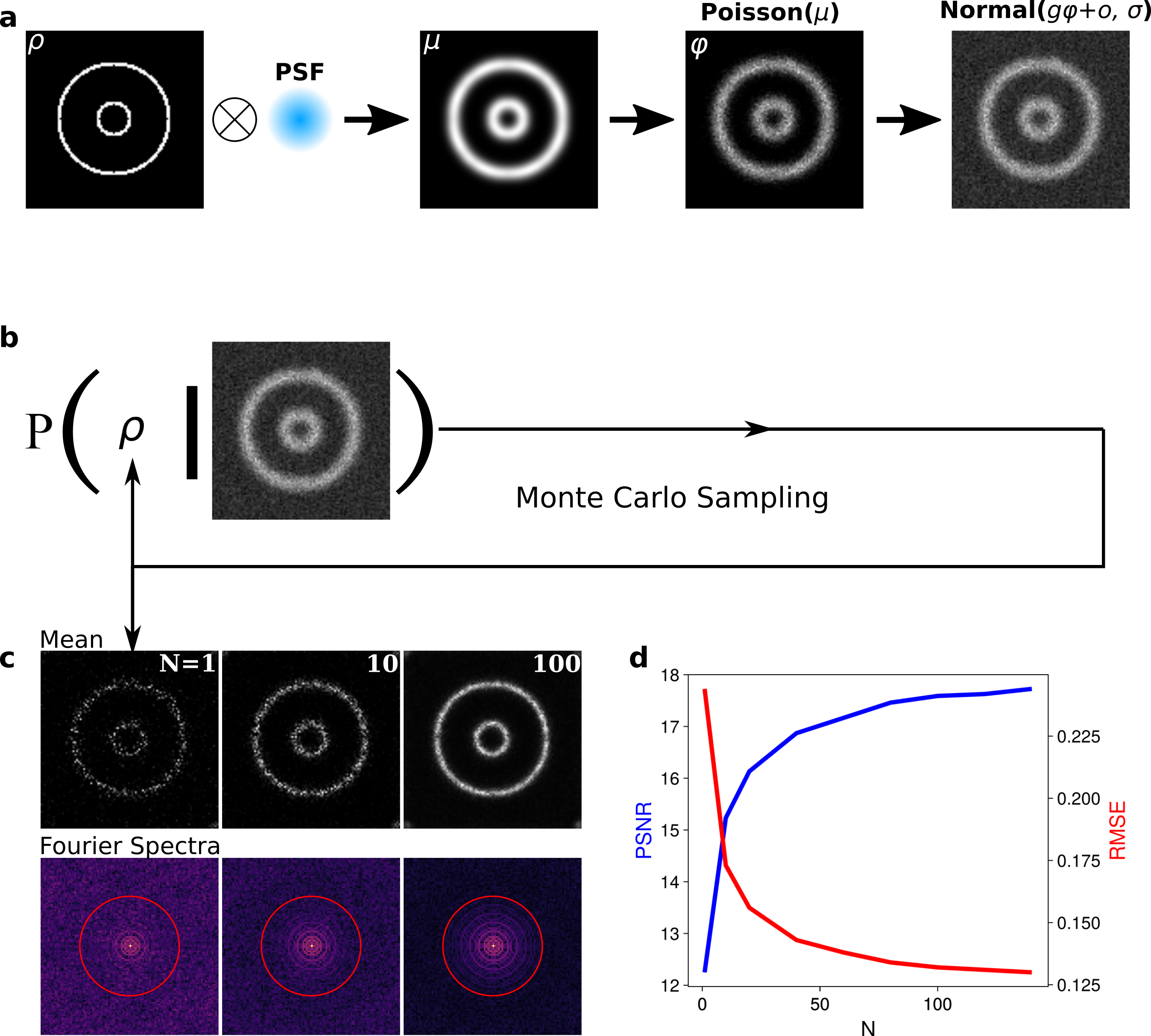}
 \caption{\textbf{Bayesian deconvolution alternative to Richardson-Lucy}. \textbf{a.} Image formation model where: 1) the imaged object $\rho$ convolved with the PSF to produce an expected convolved image $\mu$; 2) Poisson distributed photon counts $\varphi$ were replicated by corrupting the image with shot noise; and finally 3) normally distributed measurements were reproduced by further degrading with camera noise and pixelization. \textbf{b.} Inverse strategy where samples are generated from a posterior probability distribution using a Monte Carlo scheme. \textbf{c.} Mean of N Monte Carlo samples drawn from the posterior along with their Fourier spectra. Red circles represent the diffraction limit. \textbf{d.} Increasing peak sign-to-noise ratio~(PSNR) and decreasing root mean square error~(RMSE) as a function of the number of collected samples is desired and achieved by the method proposed herein.
 } 
\label{fig:imageformation}
\end{figure}

In light of these issues, we take a different approach and avoid hand-crafted regularizers or neural networks. Instead, we retain a physical image formation model as shown in Fig.~\ref{fig:imageformation} and present an unsupervised, tuning-free, and parallelizable Bayesian image deconvolution framework incorporating an accurate noise model. This framework intrinsically treats sparse and continuous samples on the same footing, and incorporates the physics of the optical setup. We mitigate the effect of overfitting from high spatial frequencies beyond the optical bandpass by placing a prior on spatial frequencies in accordance with those allowed by the optical transfer function~(OTF) and by using the mean of the Monte Carlo samples to recover a smoothed object. This physically-motivated approach allows for a robust Bayesian deconvolution redressing overfitting issues inherent to likelihood-based approaches while rigorously propagating error from noise sources.

\section{Results}
The Bayesian image deconvolution strategy we use here involves using Markov Chain Monte Carlo (MCMC) sampling to estimate a posterior probability distribution over the underlying object intensity map $\bm{\rho}$, as informed by the provided raw image. The posterior collects the data likelihood with prior knowledge codified into a prior probability distribution satisfying our desiderata. Unlike RL deconvolution, where the likelihood---a product of pixel-wise Poisson distributions---is iteratively maximized, we consider instead a likelihood integrating all known noise sources corrupting the data. Our method naturally avoids the runaway overfitting by RL not by imposing {\it ad hoc} regularization tuning or iteration cutoffs, but rather by using lack of informativeness of the data beyond the optical setup's bandpass itself to suppress the high frequencies. We achieve this by first noting that the posterior itself is naturally degenerate for high frequency modes beyond the bandpass. In other words, any one posterior sample including the maximum a posteriori~(MAP) may exhibit high-frequency artifacts. However, the mean of the collected Monte Carlo samples would smooth over those high-frequency modes while keeping low-frequency modes that are well-informed by the data intact. Consequently, the mean reflects the final deconvolved object. Furthermore, to make our sampling efficient, we devised a prior distribution that takes the OTF itself as the input and tries to suppress frequencies beyond the bandpass in the samples themselves, as described later in \nameref{Methods}. Finally, to achieve high performance, we use a parallelized MCMC inspired from Ref.~\cite{saurabh2023structured} to generate sufficient samples to draw from the posterior.  

In the following sections, we apply our strategy to simulated and experimental data demonstrating artifact-free recovery of the underlying objects, with no adjustable parameter. In contrast, RL degrades images by amplifying high-frequency noise with a rate determined by SNR -- the higher the noise in the image, the fewer iterations it takes for the noise amplification to occur.

\subsection{Simulated Data}
\begin{figure}[H]
    \centering
\includegraphics[width=0.95\linewidth]{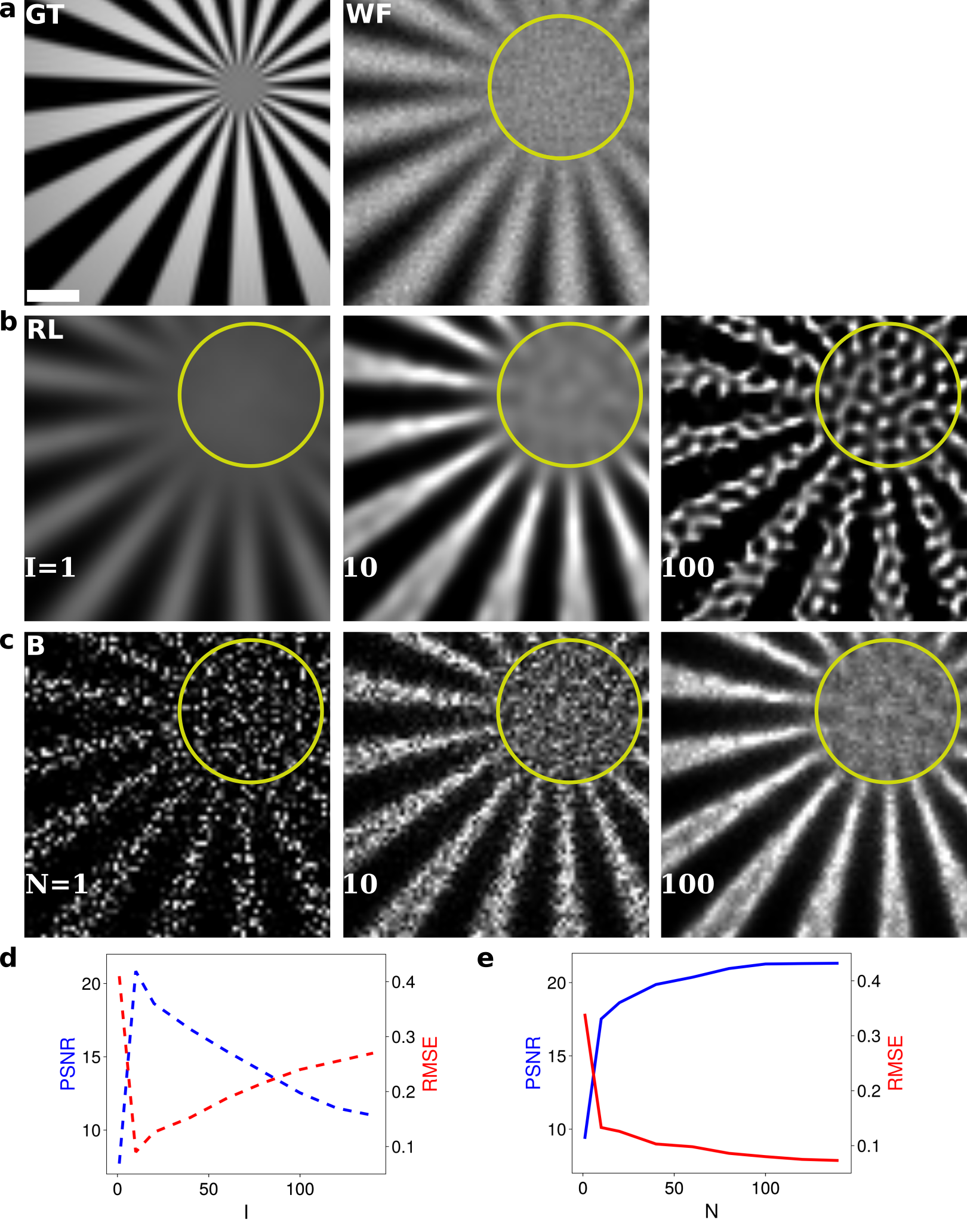}
\caption{\textbf{Deconvolution of a simulated image.} \textbf{a.} Ground truth (GT) and simulated widefield raw image (WF) assuming Poisson noise only with peak pixel photon count of 150 approximately. Yellow circles represent the diffraction limit in terms of inter-spoke spacing. The scale bar represents $0.5\upmu \mathrm{m}$. \textbf{b.} RL deconvolution~(RL) with increasing iterations~(I) from left to right. \textbf{c.} Deconvolution by our Bayesian algorithm~(B) as the number of samples~(N) used to compute the mean increases. \textbf{d-e.} PSNR and RMSE for RL (left) and Bayesian deconvolutions (right). We note optimal RL performance at around 10 iterations fo this particular sample before both PSNR and RMSE degrade motivating efforts to seek an iteration cutoff in traditional RL.
}
\label{fig:siemens}
 \end{figure}

Following the image formation model shown in Figure~\ref{fig:imageformation} and described in detail in~\nameref{Methods}, we generate a challenging image using a ground truth object with sharp edges and intensity gradients from a publicly available package, TestImages.jl~\cite{testimagesjl},  as shown in Figure~\ref{fig:siemens}.  The ground truth is convolved with, for simplicity only, a Gaussian PSF computed with a numerical aperture of $1.3$, wavelength of $510\mathrm{nm}$, and pixel size of $65\mathrm{nm}$. Notably, we only assume Poissonian photon emission for simplicity and postpone incorporating camera noise until the next section where we discuss experimental data.

In the second row of Figure~\ref{fig:siemens}, we see progressive overfitting by RL deonvolution algorithm after $100$ iterations, high-frequency, edge, striping, or ringing artifacts dominate over deconvolution, ironically visually degrading the restored image (though Figure~\ref{fig:siemens}d suggests that best peak signal-to-noise ratio~(PSNR) and root-mean-square error~(RMSE) is achieved early on at about ten iterations). On the contrary, the mean image in Bayesian deconvolution in the third row eventually converges to the diffraction-limited ground truth without giving rise to high-frequency artifacts. This stable convergence is evident from the increasing PSNR and shrinking RMSE in Figure~\ref{fig:siemens}e as we increase the number of samples used for averaging.

\subsection{Experimental Data}

\begin{figure}[H]
    \centering
\includegraphics[width=\linewidth]{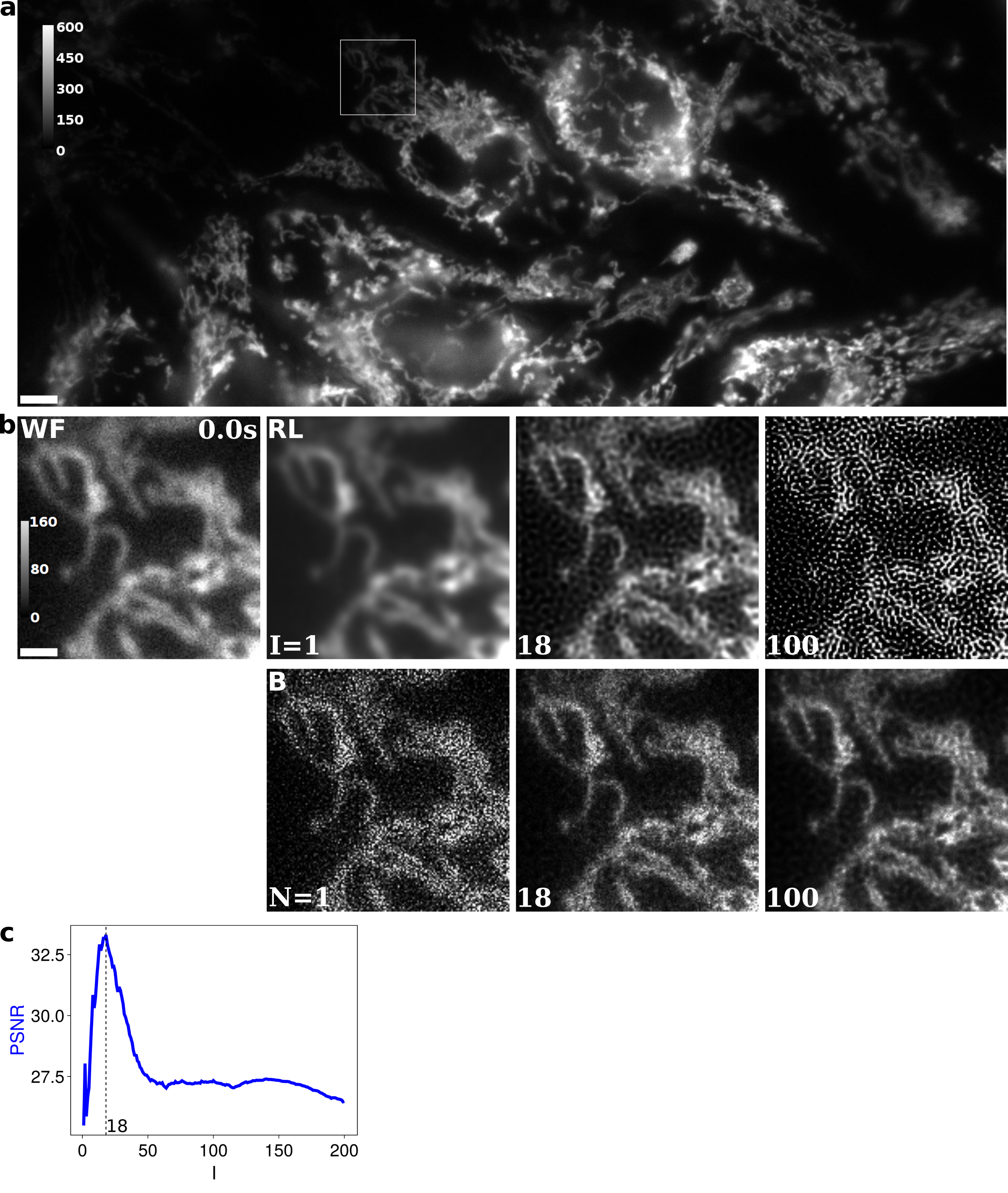}
\caption{\textbf{Deconvolution of mitochondrial network in HeLa cells labeled with Mitotracker Deep Red dye.} \textbf{a.} Raw image after subtracting average sCMOS camera offset of 100~ADU and then dividing by the average gain of 2~ADU/e$^{-}$, showing approximate photon counts. Scale bar $5.0\upmu$m.
\textbf{b.} The leftmost panel in the upper row shows the widefield raw image~(WF) from the $150\times150$ pixel inset in a. The inset lies near the edge of illumination. Scale bar $1.5\upmu$m. This image has been brightened compared to a to show structures more clearly. The remaining images in the upper row show deconvolution by RL algorithms~(RL) as the number of iterations~(I) increases. The panels in the row below show deconvolution by our Bayesian algorithm~(B) as the number of MCMC samples~(N) used to compute the mean increases. \textbf{c.} PSNR computed for the Bayesian deconvolved image at $N=100$ with RL deconvolved images by iteration I as reference, demonstrating similarity of converged Bayesian and optimal RL results.  
}
\label{fig:experimentaldata}
\end{figure}

After successfully demonstrating our Bayesian deconvolution's performance on simulated data, we now apply our method to challenging experimental images of an evolving mitochondrial network in HeLa cells, as shown in Fig.~\ref{fig:experimentaldata}. The mitochondria here are labeled with Mitotracker Deep Red dyes with \qty{660}{\nano \meter} emission peak. The light is then captured by a microscope with a numerical aperture of $1.3$ and an sCMOS camera. An Airy-disk PSF is assumed. In this image, less than $160$ photons are detected per pixel. 

Once more, as shown in Fig.~\ref{fig:experimentaldata}, RL quickly degrades the image rather than restoring it, resulting in the generation and preservation of conspicuous artifacts. Such an iterative deterioration of the deconvolved image's quality beyond $\sim 20$ iterations would typically require us to invoke at this stage a criterion for stopping iterations and whether to promote smoothness or sparsity of the final image using a regularizer. Bayesian deconvolution, on the other hand, does not amplify noise and converges to a high-contrast reconstruction by taking pixel-to-pixel camera calibration maps into account.

\noindent
Furthermore, as evident from the PSNR plot in Fig.~\ref{fig:experimentaldata}c, we find highest similarity between the Bayesian deconvolved image and RL deconvolved images in the iteration range 10-25 beyond which RL appears to amplify noise significantly.

\subsection{Uncertainty}
\begin{figure}[h]
    \centering
\includegraphics[width=0.7\linewidth]{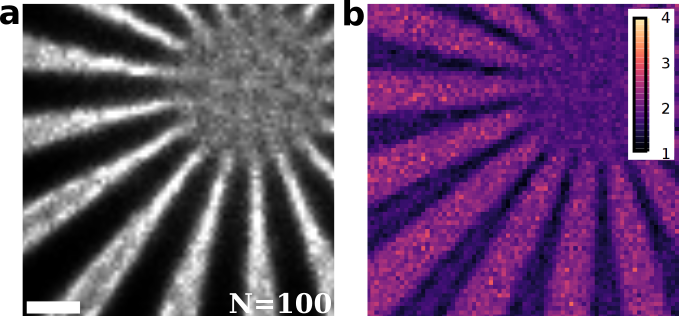}
 \caption{\textbf{Coefficient of Variation.} \textbf{a.} Restored image from Figure~\ref{fig:siemens}c obtained using Bayesian deconvolution with 100 MCMC samples. Scale bar $0.5\upmu$m. \textbf{b.} Ratio of standard deviation and mean for the Bayesian deconvolved image in the inset, providing a relative measure of uncertainty. As expected, the uncertainty is higher where information content (photon counts) is low.}
 \label{fig:uncertainty}
\end{figure}

Posterior probability distributions over the deconvolved images are estimated using MCMC sampling of the posterior, allowing us to compute not only point estimates for the restored image but also full distributions over candidate images (and their respective probability) supported by the data and, thus, by construction, uncertainty estimates as well. For example, we can use well-separated (de-correlated) MCMC samples drawn from the posterior to compute standard deviations over density estimates and other measures, such as confidence intervals. As an example, in Figure~\ref{fig:uncertainty}, we show the coefficient of variation~(CV) --- the ratio of standard deviation to the mean --- for the Bayesian deconvolved image in Figure~\ref{fig:siemens}. As expected, the relative uncertainty represented by the CV is lowest where large numbers of photons are collected. On the other hand, the solution's degeneracy (the posterior's flatness) becomes large in the darker outer regions of the image, resulting in much higher relative uncertainty.
The ability to report uncertainty stands in contrast to RL which only reports point estimates.

\subsection{Our package}
Our package for Bayesian deconvolution is available at~\cite{bdecon}. This julia code takes the raw image as input along with microscope and camera parameters. The package then runs a Monte Carlo sampler to generate hundreds of candidate deconvolved images along with their means. Furthermore, these samples can then be used to compute uncertainty estimates of choice such as quantiles and variance.

\section{Discussion}

Deconvolution is the process of denoising and removing the blurring effect of the PSF introduced by the optical setup. Numerical instabilities inherent to the convolution operator's inversion generate high-frequency artifacts that have stymied rigorous, parameter-free, numerical deconvolution. While a fully probabilistic approach is ideal, families of tools that exist for deconvolution ultimately treat the removal of these high-frequency artifacts in a user-dependent fashion or rely on expensive training of neural networks to avoid high computational cost. Here, we have successfully demonstrated parallelized artifact-free deconvolution of images with physically accurate noise modeling. Our method does not have preference for sparse or continuous solutions, unlike regularized maximum likelihood estimates. Only the data dictates whether a sparse or continuous solution is preferred. Furthermore, we imposed a prior informed by the imaging system's optics improving contrast for spatial frequencies within the diffraction limit while penalizing frequencies outside the optical setup's bandpass to improve sampling efficiency. Such a physically motivated deconvolution method, albeit more computationally demanding, eliminates the \textit{ad hoc} guesswork invoked in deciding the number of RL iterations or regularization parameters used in image restoration. 

Our framework can be generalized and improved in a number of ways. Our current implementation uses CPUs for parallelization, however, a GPU implementation can significantly reduce real computation time. Alternative camera models, such as EMCCDs~\cite{hirsch2013stochastic}, can be easily implemented into the current framework as well with the introduction of additional intermediate random variables to represent the electron multiplication stage. Finally, the physically inspired approach used to mitigate the amplification of high-frequency features immediately applies to other computational reconstruction approaches that rely on PSF of a limited size for image formation. For instance, we have successfully demonstrated a similar approach for structured illumination microscopy~\cite{saurabh2023structured} where illumination patterns multiply with the underlying fluorescent object and the parallelization of the algorithm follow the same principles as presented here. Fourier ptychography is yet another computational imaging approach where reconstructions could benefit from a fully physics approach when parallelizable.~\cite{konda2020fourier}.

\section{Methods \label{Methods}}
\subsection{Image Formation Model}
An optical system captures and projects light ({\it{i.e.}}, photons) to form an image $\boldsymbol{\mathcal{I}}$ in an image plane with positions $\bm{r}$. The system's ability to form images in this plane can be deduced from its point-source impulse response: the point spread function or $\text{PSF}(\bm{r}, \bm{r'})$. Assuming spatial invariance, the PSF describes the spatial redistribution of photon arrivals in a blurred image for a light source at $\bm{r'}$. 

\par
Therefore, the mean number of photons incident on the camera, $\bm{\mu}$, is the integral of the irradiance over the area of the $n$-th pixel, $A_n$, 
\begin{align}
    \mu_n =&  \int_{A_n} \mathrm{d}\bm{r} \int \mathrm{d}\bm{r'}~ \text{PSF}(\bm{r}, \bm{r'}) \, \rho(\bm{r'}) 
    \label{lambdafromrho}
\end{align}
which we write more compactly as 
\begin{align}
    \bm{\mu} = \int_{A_n} \mathrm{d}\bm{r} \, (\PSF \otimes \rho) (\bm{r}).\label{eq:convolution}
\end{align}
Here, $\bm{\mu}$ is the collection of expected brightness values on the n-th camera pixel, and $\otimes$ is the convolution operator. Now, the actual number of photons incident on the $n$-th  pixel is Poisson-distributed, so
\begin{align}
    \phi_n \sim& \poisson{\mu_n}.
\end{align}
Finally, the camera electronics read out the pixel value and convert the measurement to analog-to-digital units~(ADU). Assuming a CMOS camera architecture for simplicity only, the number of ADU is related to the photon number by a gain factor, $G_n$, and an offset, $o_n$. We model the effect of readout noise as a zero mean Gaussian distribution with variance $\sigma_n^2$. The final readout of the $n$-th pixel, $w_n$, is thus 
\begin{align}
    w_n  \sim& \mathbf{Normal} \left({G_n \phi_n + o_n, \sigma_{n}^2} \right).
\end{align}
\par
With this observation model for each pixel, we can now formulate the (technically completed) likelihood, $\mathcal{L}(\bm{w}, \bm{\phi} \,| \,\rho (\bm{r}),\mathbf{C} )$, for a set of observations ({\it i.e.,} a raw image) and latent variables, where $\bm{w}$ is the collection of readout values on the camera, $\bm{\phi}$ is the (latent) collection of photon counts detected by the camera, $\mathbf{C}$ is the collection of all camera parameters including gain factor, offset, and readout noise variance for each pixel, and the vertical bar ``|'' denotes conditional dependency on variables appearing on the bar's right-hand side. In cases where camera calibration maps are not available, globally constant values of gain, offset, and variance may be used.

Since $\bm{\phi}$, the number of photons incident on the camera, is typically unknown, we must marginalize (sum) over these random variables to generate the true likelihood
\begin{align} 
  \mathcal{L}(\bm{w} | \rho (\bm{r}),\mathbf{C} ) &= \sum_{\bm{\phi}} \mathcal{L}(\bm{w}, \bm{\phi} \,| \,\rho (\bm{r}),\mathbf{C} )\nonumber \\
&=  \sum_{\phi_{1:N} = 0}^{\infty} \prob{w_{1:N}, \phi_{1:N} |\, \rho (\bm{r}),   \mathbf{C}_{1:N} } \nonumber \\ 
&=\sum_{\phi_{1:N} = 0}^{\infty} \prob{w_{1:N}| \phi_{1:N} , \mathbf{C}_{1:N}} \prob{  \phi_{1:N} |  \rho(\bm{r})} ,
\end{align}
where, in the third line, we have used the chain rule for probabilities and ignored any dependency of incident photons on the camera parameters in the second term. Next, we note that expected values $\bm{\mu}^l$ for Poisson distributed photon detections are deterministically given by the convolution of Eq.~\ref{eq:convolution}. This constraint allows us to express the final likelihood as 
\begin{align}
 \mathcal{L}(\bm{w} | \rho (\bm{r}),\mathbf{C} ) & = \int \mathrm{d}\bm{\mu} \,\, \prod_{n=1}^N \left(\sum_{\phi_n = 0}^{\infty} \prob{w_{n}| \phi_{n},  \mathbf{C}_n } \prob{ \phi_{n} | \mu_{n} } \right)  \delta \left( \bm{\mu} - \int_{A_n} \mathrm{d}\bm{r} \,  \PSF \otimes \rho (\bm{r})\right) \nonumber \\
& =\int \mathrm{d}\bm{\mu} \,\,  \prod_{n=1}^N \left(\sum_{\phi_n = 0}^{\infty}  \mathbf{Normal} \left(w_n; {G_n \phi_n + o_n, \sigma_{n}^2} \right)\poisson{\phi_n; \mu_n} \right) \nonumber \\ & \hspace{6cm} \times
\delta \left( \bm{\mu} - \int_{A_n} \mathrm{d}\bm{r} \, \PSF \otimes \rho (\bm{r}) \right)  ,
\label{eq:likelihood}
\end{align}
where, assuming the independence of each camera pixel, we multiply the individual probabilities for camera readout on each pixel in order to obtain the probability over all pixels. 

We next turn to the lattice of points spanning the sample plane on which to discretize the object intensity map $\rho (\bm{r})$. For demonstration purposes, we assume that the object intensity map is such that the PSF is Nyquist sampled. We denote this object intensity map on the $m$-th point in this grid with $\rho_m$ and the collection of these values with $\bm{\rho}$.

\subsection{Prior}

Within the Bayesian paradigm, we construct the posterior probability distribution $\mathcal{P}(\bm{\rho} | \bm{w}, \mathbf{C} )$ over the underlying object intensity map $\rho$ given the image $\boldsymbol{\mathcal{I}}$.
The posterior is constructed from (and is indeed proportional to) the product of image likelihood $\mathcal{L}(\bm{w} | \bm{\rho},\mathbf{C} )$ and the prior probability distribution $\mathcal{P}(\bm{\rho})$ capturing {\it a priori} knowledge on the diffraction-limited optical setup's frequency bandpass. 

Concretely, the prior should ideally: (1) suppress frequencies beyond the bandpass (with these only revived if persistently reinforced by the data through the likelihood); (2) have the expected frequency spectrum be the bandpass spectrum itself prior to consideration of data; and (3) be degenerate with respect to the scaling of the object intensity map by a scalar constant so that it only informs features ({\it e.g.,} roughness) of the object intensity map as opposed to the amplitude.

The Dirichlet probability distribution, which is defined as 
\begin{align}
    \mathbf{Dirichlet} (\mathbf{x}; \boldsymbol{\alpha}) = \frac{1}{\text{B} (\boldsymbol{\alpha})} \prod^K_{i = 1} x_i^{\alpha_i - 1},
\end{align}
for hyperparameter $\boldsymbol{\alpha}$ elected as the normalized modulus of the optical transfer function $\left|\mathrm{OTF}(\bm{k})\right|\equiv\left|\mathcal{F}\left(\text{PSF}\right)\right|$ describing the optical setup's bandpass, satisfies all of these conditions. In the last equation, the normalization $\text{B} (\boldsymbol{\alpha})$ represents the multivariate beta function. Furthermore, the operator $\mathcal{F}$ denotes the Fourier transform. Therefore, we write the prior as
\begin{align}
\mathcal{P}(\bm{\rho}) = \int \mathrm{d}\widetilde{\bm{\rho}} \,\, \mathbf{Dirichlet} \left( \frac{|\widetilde{\bm{\rho}}|}{\sum_{\bm{k}}|\widetilde{\bm{\rho}}|} \,; \frac{|\text{OTF}|}{\sum_{\bm{k}}|\text{OTF}|} \right)  \, \delta \left( \widetilde{\bm{\rho}} - \mathcal{F}(\bm{\rho}) \right).
\end{align}
Since the Dirichlet distribution only takes vectors as inputs, we input $\widetilde{\bm{\rho}}$ and $\text{OTF}$ in their vectorized forms. Furthermore, sums in the denominators above ensure that the prior is degenerate with respect to the scaling of the object intensity map. Finally, the expectation value of the Dirichlet prior above is $\frac{|\text{OTF}|}{\sum_{\bm{k}}|\text{OTF}|}$, and therefore prior probabilities of $|\widetilde{\bm{\rho}}|$ with significant energy in frequencies above the bandpass are suppressed. Consequently, high-frequency features would have to be strongly warranted by the data (\textit{i.e.}, have high likelihoods) in order to have significant posterior probabilities.

Now, with our likelihood and prior at hand, we can finally write the posterior as 
\begin{align}
\mathcal{P}(\bm{\rho}  | \bm{w}, \mathbf{C} ) & \propto \int \mathrm{d}\bm{\mu} \,\,  \prod_{n=1}^N \left(\sum_{\phi_n = 0}^{\infty}  \mathbf{Normal} \left(w_n; {G_n \phi_n + o_n, \sigma_{n}^2} \right)\poisson{\phi_n; \mu_n} \right)  
\delta \left( \bm{\mu} -\int_{A_n} \mathrm{d}\bm{r} \,  \PSF \otimes \rho (\bm{r'}) \right) \nonumber \\ 
& \hspace{1cm} \times
\int \mathrm{d}\widetilde{\bm{\rho}} \,\, \mathbf{Dirichlet} \left( \frac{|\widetilde{\bm{\rho}}|}{\sum_{\bm{k}}|\widetilde{\bm{\rho}}|} \,; \frac{|\text{OTF}|}{\sum_{\bm{k}}|\text{OTF}|} \right)  \, \delta \left( \widetilde{\bm{\rho}} - \mathcal{F}(\bm{\rho}) \right),
\end{align}
which, imperatively, is free of tunable parameters and can be used to estimate the final deconvolved image. Here, we generate samples from the posterior using the parallelized Markov Chain Monte Carlo method adapted from Ref.~\cite{saurabh2023structured}; we then use the mean of de-correlated samples drawn from the posterior obtained upon convergence as an estimator representing the deconvolved image.

\subsection{Parallelization Technique}

\begin{figure}
\includegraphics[width=\textwidth]{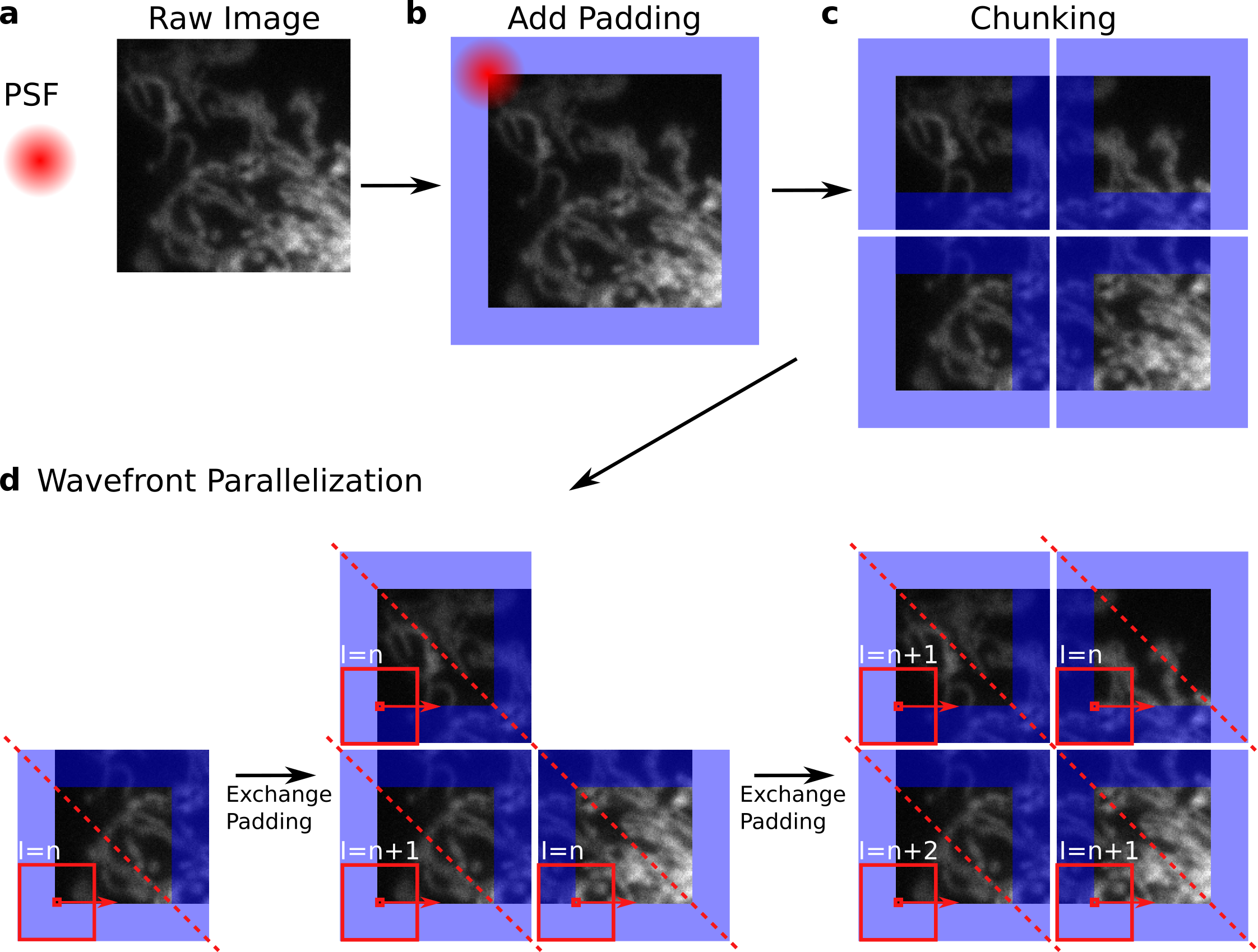}
\caption{
\textbf{Parallelized Gibbs sampling.} 
\textbf{a.} PSF and Raw Image as input for deconvolution. \textbf{b.} Padding of the size of the PSF is added to the image to perform convolutions conveniently.  \textbf{c.} Padded image is chunked. Padding in the image interior overlaps with neighboring chunks and exchanged after each MCMC iteration. \textbf{d.} At each MCMC iteration, Gibbs sampling requires sequentially updating each pixel using convolutions of the object in the red box with PSF. Each updated pixel is immediately used for updating remaining pixels during the iteration. If Gibbs sampling is initiated in a corner chunk, it can be parallelized by exchanging pixels updated at iteration $I=n$ with edge-sharing neighbor chunks and immediately moving on to the next iteration $I=n+1$ while the neighboring chunks are now being updated at iteration $I=n$. This procedure results in wavefront parallelization where all chunks are being updated simultaneously but are at different MCMC iterations.
\label{fig:parallelization}
}
\end{figure}

To make probabilistic estimates computationally inexpensive, unlike previous attempts, we do not aim to approximate the noise model. Instead, we attempt to parallelize the iterative Monte Carlo computations of the probability distributions over objects being observed by an optical setup as shown in Fig.~\ref{fig:parallelization}. Such Monte Carlo approaches involve iteratively generating randomized samples from analytically intractable probability distributions.

The first step in the parallelization emanates from the limited spread of light by diffraction. Due to the limited size of the PSF, the light entering a camera pixel originates only from a small neighborhood of the object surrounding that pixel. Or inversely said, a point on the object is only informed by a small set of camera pixels in the neighborhood of that point. Consequently, the probability distribution for a point on the object can be estimated from its immediate neighborhood. Such localization of the probability distributions allows for independent treatment of the different regions of an image and parallelization of computation. Parallelization is achieved by breaking the image into smaller equally sized chunks padded with data in the neighborhood.

An important caveat to be considered, however, is that probability distribution for a point on the object implicitly depends on knowing the contribution of the rest of the object to the light entering the camera pixels in its neighborhood. From an optimization algorithm point of view, this conditionality implies that the points on the object can only be estimated in a sequential manner once per iteration in the spirit of Gibbs sampling. Parallelization of this step is more challenging once the image has been broken into multiple chunks. A solution we have demonstrated to work successfully is wavefront parallelization where different chunks of an image, marked by an iteration wavefront, are at different Monte Carlo iterations during parallelized updates, as shown in Fig.~\ref{fig:parallelization}.

\subsection{Data Acquisition: Live HeLa cells with labeled mitochondria}
Hela cells (Kyto strain) were grown in \qty{60}{\milli \meter} glass Petri dishes on Poly-d-lysine coated \qty{40}{\milli \meter} \#1.5 coverslips (Bioptechs, 40-1313-03192) for a minimum of \qty{48}{\hour} in DMEM media (ATCC 30-2002) supplemented with \qty{10}{\percent} FBS (ATCC 30-2020) and \qty{1}{\percent} Penicillin-Streptomycin solution (ATCC, 30-2300) at \qty{37}{\celsius} and \qty{5}{\percent} CO2. Cells were live stained with \qty{200}{\nano \molar} Mitotracker Deep Red (ThermoFisher, M22426) in DMEM medium for \qty{15}{\minute} in the same incubation environment. The staining solution was then aspirated off, and fresh DMEM medium was added for \qty{5}{\minute} to rinse. The sample coverslip was then transferred to an open-top Bioptechs FCS2 chamber and imaged in pre-warmed (\qty{37}{\celsius}) DMEM culture medium.

HeLa cells with labeled mitochondria were imaged on the same instrument described for the Argo-SIM calibration slide using pseudo-widefield mode for \qty{635}{\nano \meter} excitation light derived from a Lasever diode laser (LSR635-500). The camera integration time was fixed at \qty{100}{\milli \second}, and the signal level was varied by changing the illumination time using the DMD as a fast shutter. Illumination times were \qty{100}{\milli \second}, \qty{10}{\milli \second}, and \qty{1}{\milli \second}.

\subsection{Richardson-Lucy Deconvolution}
We used a Julia package, DeconvOptim.jl~\cite{Wechsler2023}, and a python package, Scikit-Image~\cite{scikit}, to test all Richardson-Lucy deconvolutions.

\section{Data Availability}

Simulation data, experimental data, and the Julia code for Bayesian deconvolution are available at~\cite{bdecon}.

\section{Author Contributions}
SP and AS worked on the first derivation of the prior. ZHH developed the initial serial implementation in Python, and AS developed the final parallelized Julia code. PTB and DPS provided multiple insights on all aspects of the project and datasets for the mitochondria networks in HeLa cells. ZHH and AS wrote the first draft of the paper and everyone subsequently participated in editing. SP oversaw all aspects of the project.

\section{Acknowledgments}
First and foremost we thank Andrew York (Calico) for inspiring and encouraging us to think deeply about the limitations of RL. Without these discussions, we would never have considered applying Bayesian tools to RL. We also thank Maxwell Schweiger for helpful discussions along the way pertaining to the form of the prior. PTB and DPS acknowledge funding support from the NIH (RF1MH128867) and Scialog, Research Corporation for Science Advancement, and Frederick Gardner Cottrell Foundation (28041). SP acknowledges support from the NIH (R01GM134426, R01GM130745, and R35GM148237).

\newpage

\bibliographystyle{unsrt}
\bibliography{Bibliography}

\end{document}